\documentclass[10pt,twocolumn,letterpaper]{article}

\usepackage{cvpr}              

\usepackage[dvipsnames, svgnames, x11names]{xcolor}
\usepackage{graphicx}
\usepackage{amsmath}
\usepackage{amssymb}
\usepackage{booktabs}
\usepackage{multirow}
\usepackage{makecell}
\usepackage{colortbl}

\usepackage[pagebackref,breaklinks,colorlinks]{hyperref}

\usepackage[capitalize]{cleveref}
\crefname{section}{Sec.}{Secs.}
\Crefname{section}{Section}{Sections}
\Crefname{table}{Table}{Tables}
\crefname{table}{Tab.}{Tabs.}

\def\cvprPaperID{8034} 
\def\confName{CVPR}
\def\confYear{2023}

\begin{document}

\title{Lite-Mono: A Lightweight CNN and Transformer Architecture for Self-Supervised Monocular Depth Estimation}

\author{Ning Zhang\thanks{Corresponding Author}\qquad Francesco Nex\qquad George Vosselman\qquad Norman Kerle \\
 University of Twente\\
{\tt\small \{n.zhang, f.nex, george.vosselman, n.kerle\}@utwente.nl}
}
\maketitle

\begin{abstract}
   Self-supervised monocular depth estimation that does not require ground truth for training has attracted attention in recent years. It is of high interest to design lightweight but effective models so that they can be deployed on edge devices. Many existing architectures benefit from using heavier backbones at the expense of model sizes. This paper achieves comparable results with a lightweight architecture. Specifically, the efficient combination of CNNs and Transformers is investigated, and a hybrid architecture called Lite-Mono is presented. A Consecutive Dilated Convolutions (CDC) module and a Local-Global Features Interaction (LGFI) module are proposed. The former is used to extract rich multi-scale local features, and the latter takes advantage of the self-attention mechanism to encode long-range global information into the features. Experiments demonstrate that Lite-Mono outperforms Monodepth2 by a large margin in accuracy, with about 80\% fewer trainable parameters. Our codes and models are available at \url{https://github.com/noahzn/Lite-Mono}.
\end{abstract}


\section{Introduction}
\label{sec:intro}

Many applications in the field of robotics, autonomous driving, and augmented reality rely on depth maps, which represent the 3D geometry of a scene. Since depth sensors increase costs, research on inferring depth maps using Convolutional Neural Networks (CNNs) from images emerged. With the annotated depth one can train a regression CNN to predict the depth value of each pixel on a single image~\cite{1,2,3}. Lacking large-scale accurate dense ground-truth depth for supervised learning, self-supervised methods that seek supervisory signals from stereo-pairs of frames or monocular videos are favorable and have made great progress in recent years. These methods regard the depth estimation task as a novel view synthesis problem and minimize an image reconstruction loss~\cite{5,6,7,8,9}. The camera motion is known when using stereo-pairs of images, so a single depth estimation network is adopted to predict depth. But if only using monocular videos for training an additional pose network is needed to estimate the motion of the camera. Despite this, self-supervised methods that only require monocular videos are preferred, as collecting stereo data needs complicated configurations and data processing. Therefore, this paper also focuses on monocular video training. 

\begin{figure}[!t]
  \centering
\includegraphics[width=1\linewidth]{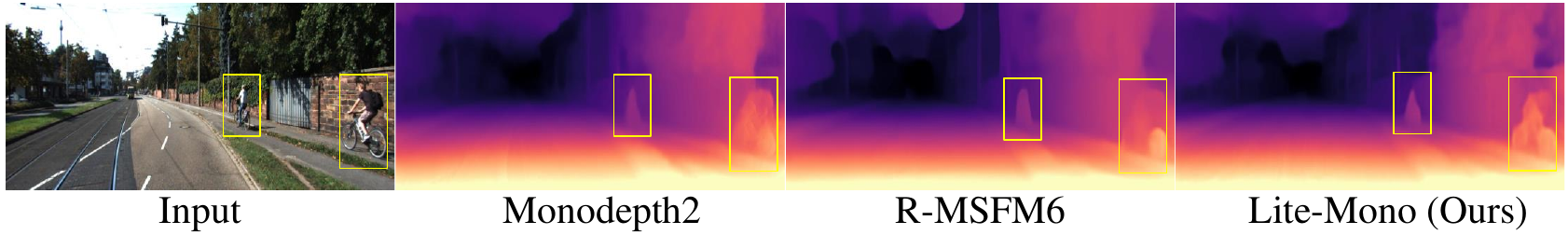}

   \caption{The proposed Lite-Mono has fewer parameters than Monodepth2~\cite{9} and R-MSFM~\cite{10}, but generates more accurate depth maps.}
   \label{fig:teaser}
   \vspace{-5mm}
\end{figure}

In addition to increasing the accuracy of monocular training by introducing improved loss functions~\cite{9} and semantic information~\cite{8,26} to mitigate the occlusion and moving objects problems, many works focused on designing more effective CNN architectures~\cite{7,10,11,12,13}. However, the convolution operation in CNNs has a local receptive field, which cannot capture long-range global information. To achieve better results a CNN-based model can use a deeper backbone or a more complicated architecture~\cite{9, 14, 27}, which also results in a larger model size. The recently introduced Vision Transformer (ViT)~\cite{15} is able to model global contexts, and some recent works apply it to monocular depth estimation architectures~\cite{16, 18} to obtain better results. However, the expensive calculation of the Multi-Head Self-Attention (MHSA) module in a Transformer hinders the design of lightweight and fast inference models, compared with CNN models~\cite{16}. 

This paper pursues a lightweight and efficient self-supervised monocular depth estimation model with a hybrid CNN and Transformer architecture. In each stage of the proposed encoder a Consecutive Dilated Convolutions (CDC) module is adopted to capture enhanced multi-scale local features. Then, a Local-Global Features Interaction (LGFI) module is used to calculate the MHSA and encode global contexts into the features. To reduce the computational complexity the cross-covariance attention~\cite{21} is calculated in the channel dimension instead of the spatial dimension. The contributions of this paper can be summarized in three aspects.
\begin{itemize}
\item[$\bullet$] A new lightweight architecture, dubbed Lite-Mono, for self-supervised monocular depth estimation, is proposed. Its effectiveness with regard to the model size and FLOPs is demonstrated.
\item[$\bullet$] The proposed architecture shows superior accuracy on the KITTI~\cite{29} dataset compared with competitive larger models. It achieves state-of-the-art with the least trainable parameters. The model's generalization ability is further validated on the Make3D~\cite{31} dataset. Additional ablation experiments are conducted to verify the effectiveness of different design choices.

\item[$\bullet$] The inference time of the proposed method is tested on an NVIDIA TITAN Xp and a Jetson Xavier platform, which demonstrates its good trade-off between model complexity and inference speed. 

\end{itemize}

The remainder of the paper is organized as follows. Section~\ref{sec:relatedwork} reviews some related research work. Section~\ref{sec:lite-mono} illustrates the proposed method in detail. Section~\ref{sec:exp} elaborates on the experimental results and discussion. Section~\ref{sec:conclusions} concludes the paper.

\section{Related work}
\label{sec:relatedwork}

\subsection{Monocular depth estimation using deep learning}
Single image depth estimation is an ill-posed problem, because a 2D image may correspond to many 3D scenes at different scales. Methods using deep learning can be roughly divided into two categories. 

\textbf{Supervised depth estimation.} Using ground-truth depth maps as supervision, a supervised deep learning network is able to extract features from input images and learn the relationship between depth and RGB values. Eigen \textit{et al.}~\cite{1} first used deep networks to estimate depth maps from single images. They designed a multi-scale network to combine global coarse depth maps and local fine depth maps. Subsequent works introduced some post-processing techniques, such as Conditional Random Fields (CRF), to improve the accuracy~\cite{42,43,44}. Laina~\textit{et al.}~\cite{2} proposed to use a new up-sampling module and the reverse Huber loss to improve the training. Fu~\textit{et al.}~\cite{3} adopted a multi-scale network, and treated the depth estimation as an ordinal regression task. Their method achieved higher accuracy and faster convergence.

~\textbf{Self-supervised depth estimation.} Considering that large-scale annotated datasets are not always available, self-supervised depth estimation methods that do not require ground truth for training have attracted some attention. Garg~\textit{et al.}~\cite{4} regarded the depth estimation as a novel view synthesis problem, and proposed to minimize a photometric loss between an input left image and the synthesized right image. Their method was self-supervised, as the supervisory signal came from the input stereo pairs. Godard~\textit{et al.}~\cite{5} extended this work and achieved higher accuracy by introducing a left-right disparity consistency loss. Apart from using stereo pairs the supervisory signal can also come from monocular video frames. Zhou~\textit{et al.}~\cite{6} trained a separate multi-view pose network to estimate the pose between two sequential frames. To improve the robustness when dealing with occlusion and moving objects they also used an explainability prediction network to ignore target pixels that violate view synthesis assumptions. To model dynamic scenes other works introduced multi-task learning, such as optical flow estimation~\cite{7} and semantic segmentation~\cite{8,40}, or introduced additional constraints, such as uncertainty estimation~\cite{39, 41}. Godard~\textit{et al.}~\cite{9} found that without introducing extra learning tasks they could achieve competitive results by simply improving the loss functions. They proposed Monodepth2, which used a minimum reprojection loss to mitigate occlusion problems, and an auto-masking loss to filter out moving objects that have the same velocity as the camera. This work is also based on their self-supervised training strategy.

\begin{figure*}[!ht]
\begin{center}
\includegraphics[width=14cm]{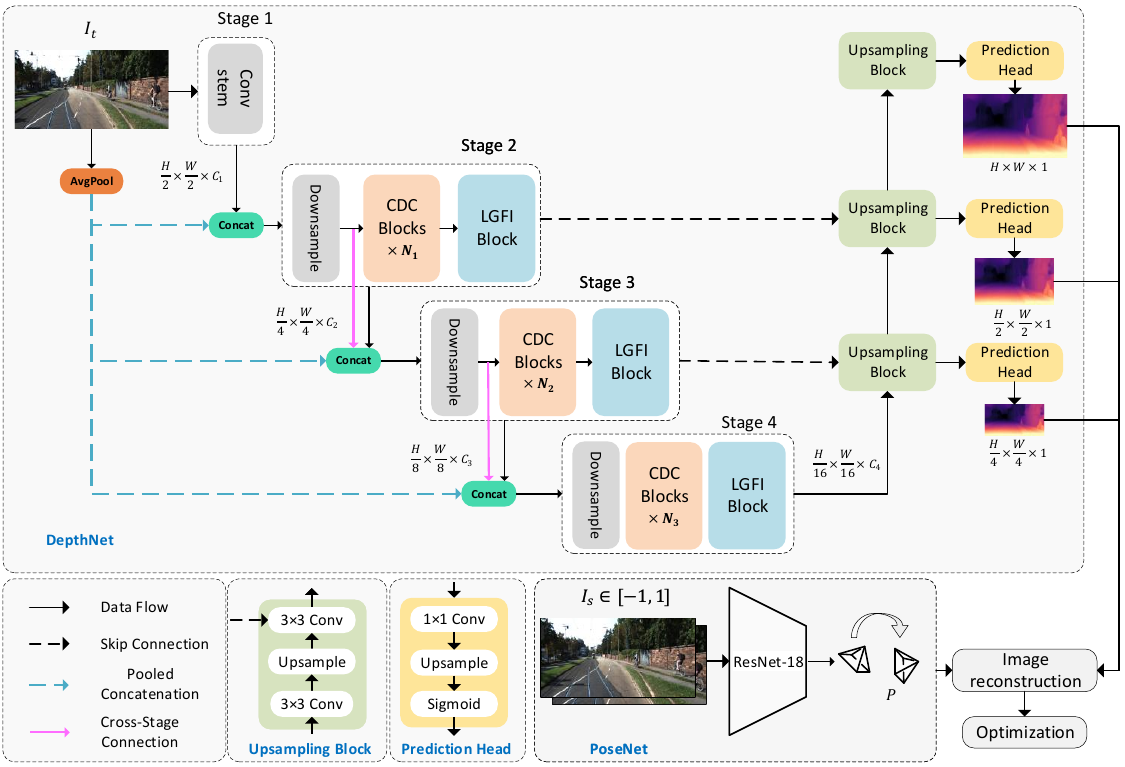}
\end{center}
 \vspace{-3mm}
\caption{~\textbf{Overview of the proposed Lite-Mono.} Lite-Mono has an encoder-decoder DepthNet for depth prediction, and a commonly used PoseNet~\cite{9, 10} to estimate poses between adjacent monocular frames. The encoder of the DepthNet consists of four stages, and it uses Consecutive Dilated Convolutions (CDC) modules and Local-Global Features Interaction (LGFI) modules to extract rich hierarchical features. The details of these modules are shown in Figure~\ref{fig:modules}.}
\label{fig:network}
\end{figure*}
\subsection{Advanced architectures for depth estimation}
Network architectures also play an important role in achieving good results in monocular depth estimation. By replacing the network architecture from the VGG model~\cite{11} with a ResNet~\cite{12} Yin~\textit{et al.}~\cite{7} achieved better results. Yan~\textit{et al.}~\cite{13} used a channel-wise attention module to capture long-range multi-level information and enhance local features. Zhou~\textit{et al.}~\cite{14} also used an attention module to obtain a better feature fusion. Zhao~\textit{et al.}~\cite{10} proposed a small architecture that used a feature modulation module to learn multi-scale features, and demonstrated the method's superiority. To reduce model parameters they only used the first three stages of ResNet18~\cite{12} as the backbone. With the rise of vision transformer (ViT)~\cite{15} recent work applied it to various computer vision tasks~\cite{17,32,33,34,35}, and achieved promising results. However, research incorporating Transformers in depth estimation architectures is still limited. Varma~\textit{et al.}~\cite{16} adopted the Dense Prediction Transformer~\cite{17} for self-supervised monocular depth estimation, and added another prediction head to estimate the camera's intrinsic. Bae~\textit{et al.}~\cite{18} proposed a hybrid architecture of CNN and Transformer that enhanced CNN features by Transformers. However, due to the high computational complexity of Multi-Head Self-Attention (MHSA) in a ViT, the above-mentioned Transformer-based methods have more trainable parameters and have a large speed gap compared with methods only using CNNs~\cite{16}. MonoViT~\cite{46} uses MPViT~\cite{50} as its encoder and has achieved state-of-the-art accuracy. Nevertheless, the use of multiple parallel blocks in MonoViT slows down its speed.

\section{The proposed framework: Lite-Mono}
\label{sec:lite-mono}
\subsection{Design motivation and choices}
Several papers demonstrated that a good encoder can extract more effective features, thus improving the final result~\cite{9, 12, 14}. This paper focuses on designing a lightweight encoder that can encode effective features from the input images. Figure~\ref{fig:network} shows the proposed architecture. It consists of an encoder-decoder DepthNet (Section ~\ref{sec:depthnet}) and a PoseNet (Section ~\ref{sec:posenet}). The DepthNet estimates multi-scale inverse depth maps of the input image, and the PoseNet estimates the camera motion between two adjacent frames. Then, a reconstructed target image is generated, and the loss is computed to optimize the model (Section ~\ref{sec:ssl}).

\textbf{Enhanced local features.}
Using shallow instead of deeper networks can effectively reduce the size of a model. As mentioned shallow CNNs have very limited receptive fields, while using dilated convolution~\cite{37} is helpful to enlarge receptive fields. By stacking the proposed Consecutive Dilated Convolutions (CDC) the network is able to "observe" the input at a larger area, while not introducing extra training parameters.

\textbf{Low-computation global information.} The enhanced local features are not enough to learn a global representation of the input without the help of Transformers to model long-range information. The MHSA module in the original Transformer~\cite{15} has a linear computational complexity to the input dimension, hence it limits the design of lightweight models. Instead of computing the attention across the spatial dimension the proposed Local-Global Features Interaction (LGFI) module adopts the cross-covariance attention~\cite{21} to compute the attention along the feature channels. Comparing with the original self-attention~\cite{15} it reduces the memory complexity from $ \mathcal{O}(hN^2+Nd)$ to $ \mathcal{O}(d^2/h+Nd)$, and reduces the time complexity from $ \mathcal{O}(N^2d)$ to $ \mathcal{O}(Nd^2/h)$, where $h$ is the number of attention heads. The proposed architecture is described in detail below.

\begin{table*}[!thb]
\begin{center}
\scalebox{0.76}{
\begin{tabular}{cccccc}
\hline
Output Size&Layers  & Lite-Mono-tiny & Lite-Mono-small & Lite-Mono & Lite-Mono-8M\\

\hline\hline
$640\times 192$&Input\\
\hline
\multirow2*{$320\times 96$}&\multirow2*{Conv Stem}&$3\times 3, 32, stride=2$&$3\times 3, 48, stride=2$&$3\times 3, 48, stride=2$&$3\times 3, 64, stride=2$\\
&&$\left[3\times 3, 32\right]\times 2$&$\left[3\times 3, 48\right]\times 2$&$\left[3\times 3, 48\right]\times 2$&$\left[3\times 3, 64\right]\times 2$\\
\hline
$160\times 48$&Downsampling&$3\times 3, 32, stride=2$&$3\times 3, 48, stride=2$&$3\times 3, 48, stride=2$&$3\times 3, 64, stride=2$\\
\multirow{2}*{Stage 1}&CDC blocks& $\left[3\times 3, 32\right]\times 3$&$\left[3\times 3, 48\right]\times 3$&$\left[3\times 3, 48\right]\times 3$&$\left[3\times 3, 64\right]\times 3$\\
&LGFI block&$dilation=1, 2, 3$&$dilation=1, 2, 3$&$dilation=1, 2, 3$&$dilation=1, 2, 3$\\

\hline

$80\times 24$&Downsampling&$3\times 3, 64, stride=2$&$3\times 3, 80, stride=2$&$3\times 3, 80, stride=2$&$3\times 3, 128, stride=2$\\
\multirow{2}*{Stage 2}&CDC blocks& $\left[3\times 3, 64\right]\times 3$&$\left[3\times 3, 80\right]\times 3$&$\left[3\times 3, 80\right]\times 3$&$\left[3\times 3, 128\right]\times 3$\\
&LGFI block&$dilation=1, 2, 3$&$dilation=1, 2, 3$&$dilation=1, 2, 3$&$dilation=1, 2, 3$\\

\hline

$40\times 12$&Downsampling&$3\times 3, 128, stride=2$&$3\times 3, 128, stride=2$&$3\times 3, 128, stride=2$&$3\times 3, 224, stride=2$\\

\multirow{2}*{Stage 3}&CDC blocks& $\left[3\times 3, 128\right]\times 6$&$\left[3\times 3, 128\right]\times 6$&$\left[3\times 3, 128\right]\times 9$&$\left[3\times 3, 224\right]\times 9$\\
&LGFI block&$dilation=1, 2, 3, 2, 4, 6$&$dilation=1, 2, 3, 2, 4, 6$&$dilation=\left[1, 2, 3\right]\times 2, 2, 4, 6$&$dilation=\left[1, 2, 3\right]\times 2, 2, 4, 6$\\

\hline\hline
\#Params. (M)&&2.0&2.3&2.9&8.1\\
\hline

\end{tabular}}
\end{center}
\vspace{-3mm}
\caption{\textbf{Four variants of the proposed depth encoder.} $[3\times3,C]\times N$ means that a CDC block uses the $3\times 3$ kernel size to output $C$ channels, and repeats for $N$ times. The dilation rate used in each CDC block is also listed.}
\label{table:variants}
\vspace{-3mm}
\end{table*}

\begin{figure}[!htb]
\begin{center}
\includegraphics[width=0.8\linewidth]{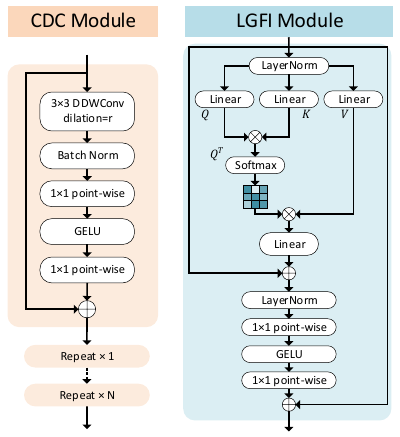}
\end{center}
 \vspace{-4mm}
\caption{\textbf{Structures of the proposed Consecutive Dilated Convolutions (CDC) module and Local-Global Features Interaction (LGFI) module.} In each stage the CDC module with different dilation rate is repeated for $N$ times.}
\label{fig:modules}
\end{figure}

\subsection{DepthNet}
\label{sec:depthnet}
\textbf{Depth encoder.} 
The proposed Lite-Mono aggregates multi-scale features across four stages. The input image with size $H\times W\times 3$ is first fed into a convolution stem, where the image is down-sampled by a $3\times 3$ convolution. Following two additional $3\times 3$ convolutions with $stride=1$ for local feature extraction, the feature maps of size $\frac{H}{2}\times \frac{W}{2}\times C_1$ are obtained. In the second stage the features are concatenated with the pooled three-channel input image, and another $3\times 3$ convolution with $stride=2$ is adopted to down-sample the feature maps, resulting in feature maps with size $\frac{H}{4}\times \frac{W}{4}\times C_2$. Concatenating features with the average-pooled input image in a down-sampling layer can reduce the spatial information loss caused by the reduction of feature size, which is inspired by ESPNetv2~\cite{18}. Then, the proposed Consecutive Dilated Convolutions (CDC) module and the Local-Global Features Interaction (LGFI) module learn rich hierarchical feature representations. The down-sampling layers in the second and the third stage also receive the concatenated features output by the previous down-sampling layer. This design is similar to the residual connection proposed in ResNet~\cite{12}, and is able to model better cross-stage correlation. Similarly, the output feature maps are further fed into the third and the fourth stages, and output features of dimension $\frac{H}{8}\times \frac{W}{8}\times C_3$ and $\frac{H}{16}\times \frac{W}{16}\times C_4$.

\textbf{Consecutive Dilated Convolutions (CDC).}
The proposed CDC module utilizes dilated convolutions to extract multi-scale local features. Different from using a parallel dilated convolution module only in the last layer of the network~\cite{20} we insert several consecutive dilated convolutions with different dilation rates into each stage for adequate multi-scale contexts aggregation. 

Given a two-dimensional signal $x[i]$ the output $y[i]$ of a 2D dilated convolution can be defined as:

\begin{equation}
  y[i] = \sum_{k=1}^{K}x[i+r\cdot k]w[k],
  \label{eq:2d dilated}
\end{equation}
where $w[k]$ is a filter with length $K$, and $r$ denotes the dilation rate used to convolve the input $x[i]$. In a standard non-dilated convolution $r=1$. By using a dilated convolution the network can keep the size of the output feature map fixed while achieving a larger receptive field. Considering an input feature $X$ with dimension $H\times W\times C$ our CDC module outputs $\hat{X}$ as follows:
\begin{equation}
  \hat{X} = 
  X+Linear_G(Linear(BN(DDWConv_r(X)))),
  \label{eq:CDC}
\end{equation}
where $Linear_G$ denotes a point-wise convolution operation, followed by the $GELU$~\cite{45} activation. $BN$ is a batch normalization layer, and $DDWConv_r(\cdot)$ is a $3\times 3$ depth-wise dilated convolution with dilation rate $r$.

\textbf{Local-Global Features Interaction (LGFI).}
 Given an input feature map $X$ with dimension $H\times W\times C$ it is linearly projected to the same dimensional queries $Q=XW_q$, keys $K=XW_k$, and values $V=XW_v$, where $W_q$, $W_k$, and $W_v$ are weight matrices. The cross-covariance attention~\cite{21} is used to enhance the input $X$:

\begin{equation}
  \widetilde{X} = Attention(Q,K,V) + X,
  \label{eq:lmhsa}
\end{equation}
where $Attention(Q,K,V)=V\cdot Softmax(Q^T\cdot K)$. Then, the non-linearity of the features can be increased:
\begin{equation}
  \hat{X} = X+ Linear_G(Linear(LN(\widetilde{X}))),
  \label{eq:ffn}
\end{equation}
where $LN$ is a layer normalization~\cite{38} operation. According to different channel numbers, CDC blocks, and dilation rates, four variants of the depth encoder are designed. Table~\ref{table:variants} shows more details.

\textbf{Depth decoder.}
Different from using a complicated up-sampling method~\cite{10} or introducing additional attention modules~\cite{18} Lite-Mono uses a depth decoder adapted from~\cite{9}. As shown in Figure~\ref{fig:network} it increases the spatial dimension using bi-linear up-sampling, and uses convolutional layers to concatenate features from three stages of the encoder. Each up-sampling block follows a prediction head to output the inverse depth map at full, $\frac{1}{2}$, and $\frac{1}{4}$ resolution, respectively.

\begin{table*}[ht!]
\begin{center}
\scalebox{0.7}{
\begin{tabular}{ccc|cccc|ccc|c}
\hline
\multirow{2}*{Method}& \multirow{2}*{Year}&
\multirow{2}*{Data}&
\multicolumn{4}{c|}{Depth Error ($\downarrow$)}&
\multicolumn{3}{c|}{Depth Accuracy ($\uparrow$)}&
\multicolumn{1}{c}{Model Size ($\downarrow$)}\\

\cline{4-11}
&&& Abs Rel & Sq Rel & RMSE & RMSE log &$\delta <1.25$ &$\delta <1.25^2$&$\delta <1.25^3$&Params. \\
\cline{1-11}
GeoNet~\cite{7}&2018&M&0.149&1.060&5.567&0.226&0.796&0.935&0.975&31.6M\\
DDVO~\cite{25}&2018&M&0.151&1.257&5.583&0.228&0.810&0.936&0.974&28.1M\\
Monodepth2-Res18~\cite{9}&2019&M&0.115&0.903&4.863&0.193&0.877&0.959&0.981&14.3M\\
Monodepth2-Res50~\cite{9}&2019&M&0.110&0.831&4.642&0.187&0.883&\underline{0.962}&\underline{0.982}&32.5M\\
SGDepth~\cite{26}&2020&M+Se&0.113&0.835&4.693&0.191&0.879&0.961&0.981&16.3M\\
Johnston~\textit{et al.}~\cite{28}&2020&M&0.111&0.941&4.817&0.189&\underline{0.885}&0.961&0.981&14.3M+\\
CADepth-Res18~\cite{13}&2021&M&0.110&0.812&4.686&0.187&0.882&\underline{0.962}&\textbf{0.983}&18.8M\\
HR-Depth~\cite{27}&2021&M&0.109&\underline{0.792}&4.632&0.185&0.884&\underline{0.962}&\textbf{0.983}&14.7M\\
Lite-HR-Depth~\cite{27}&2021&M&0.116&0.845&4.841&0.190&0.866&0.957&\underline{0.982}&3.1M\\
R-MSFM3~\cite{10}&2021&M&0.114&0.815&4.712&0.193&0.876&0.959&0.981&3.5M\\
R-MSFM6~\cite{10}&2021&M&0.112&0.806&4.704&0.191&0.878&0.960&0.981&3.8M\\
MonoFormer~\cite{18}&2022&M&\underline{0.108}&0.806&\underline{4.594}&\underline{0.184}&0.884&\textbf{0.963}&\textbf{0.983}&23.9M+\\

\textbf{Lite-Mono-tiny (Ours)}&2023&M&0.110&0.837&4.710&0.187&0.880&0.960&\underline{0.982}&\textbf{2.2M}\\
\textbf{Lite-Mono-small (Ours)}&2023&M&   0.110  &   0.802  &   4.671  &   0.186  &   0.879  &   0.961  &   \underline{0.982}&\underline{2.5M}  \\
\textbf{Lite-Mono (Ours)}&2023&M&\textbf{0.107}  &   \textbf{0.765}  &   \textbf{4.561}  &   \textbf{0.183}  &   \textbf{0.886}  &   \textbf{0.963}  &   \textbf{0.983}  &
3.1M \\

\cline{1-11}
\hline

Monodepth2-Res18~\cite{9}&2019&M\dag&0.132&1.044&5.142&0.210&0.845&0.948&0.977&14.3M\\
Monodepth2-Res50~\cite{9}&2019&M\dag&0.131&1.023&5.064&0.206&0.849&0.951&\underline{0.979}&32.5M\\
R-MSFM3~\cite{10}&2021&M\dag&0.128&0.965&5.019&0.207&0.853&0.951&0.977&3.5M\\
R-MSFM6~\cite{10}&2021&M\dag&0.126&0.944&4.981&0.204&\underline{0.857}&\underline{0.952}&0.978&3.8M\\

\textbf{Lite-Mono-tiny (Ours)}&2023&M\dag&0.125&0.935&4.986&0.204&0.853&0.950&0.978&\textbf{2.2M}\\
\textbf{Lite-Mono-small (Ours)}&2023&M\dag&   \underline{0.123}  &   \underline{0.919}  &   \underline{4.926}  &   \underline{0.202}  &   \textbf{0.859}  &   0.951  &   0.977&\underline{2.5M}  \\
\textbf{Lite-Mono (Ours)}&2023&M\dag&\textbf{0.121}  &   \textbf{0.876}  &   \textbf{4.918}  &   \textbf{0.199}  &   \textbf{0.859}  &   \textbf{0.953}  &   \textbf{0.980}&3.1M \\
\hline

Monodepth2-Res18~\cite{9} &2019&M*&0.115&0.882&4.701&0.190&0.879&0.961&0.982&14.3M\\
R-MSFM3~\cite{10}&2021&M*&0.112&0.773&4.581&0.189&0.879&0.960&0.982&3.5M\\
R-MSFM6~\cite{10}&2021&M*&0.108&0.748&4.470&0.185&0.889&0.963&0.982&3.8M\\
HR-Depth~\cite{27}&2021&M*&0.106&0.755&   4.472&0.181&0.892&  \underline{0.966}&\textbf{0.984}&14.7M\\

\textbf{Lite-Mono-tiny (Ours)}&2023&M*&0.104  &   0.764  &   4.487  &   0.180  &   0.892  &   0.964  &   \underline{0.983}&\textbf{2.2M}\\
\textbf{Lite-Mono-small (Ours)}&2023&M*&0.103  &   0.757  &   4.449  &   0.180  &  0.894  &   0.964  &   \underline{0.983}&\underline{2.5M}\\
\textbf{Lite-Mono (Ours)}&2023&M*& \underline{0.102}  &   \underline{0.746}  &   \underline{4.444}  &   \underline{0.179}  &   \underline{0.896}  &   0.965  &   \underline{0.983}&3.1M  \\
\textbf{Lite-Mono-8M (Ours)}&2023&M*&\textbf{0.097}  &   \textbf{0.710}  &   \textbf{4.309}  &   \textbf{0.174}  &   \textbf{0.905}  &   \textbf{0.967}  &   \textbf{0.984}&8.7M\\

\hline
\hline


MonoViT-tiny~\cite{46}&2022&M&0.102&0.733&4.459&\textbf{0.177}&0.895&\textbf{0.965}&\textbf{0.984}&10.3M\\
\textbf{Lite-Mono-8M (Ours)}& 2023&M&   \textbf{0.101}  &   \textbf{0.729}  &   \textbf{4.454}  &   0.178  &   \textbf{0.897}  &   \textbf{0.965}  &   0.983&\textbf{8.7M}  \\

\hline

\end{tabular}}
\end{center}
\vspace{-3mm}
\caption{\textbf{Comparison of Lite-Mono with some recent representative methods on the KITTI benchmark using the Eigen split~\cite{30}.} All input images are resized to $640\times 192$ unless otherwise specified. The best and the second best results are highlighted in \textbf{bold} and \underline{underlined}, respectively. "M": KITTI monocular videos, "M+Se": monocular videos + semantic segmentation, "M*": input resolution $1024\times 320$, "M$\dag$": without pre-training on ImageNet~\cite{36}.}
\label{exp:full}
\end{table*}

\subsection{PoseNet}
\label{sec:posenet}
Following~\cite{9, 10} this paper uses the same PoseNet for pose estimation. To be specific, a pre-trained ResNet18 is used as the pose encoder, and it receives a pair of color images as input. A pose decoder with four convolutional layers is used to estimate the corresponding 6-DoF relative pose between adjacent images.

\begin{figure*}[!ht]
  \centering
   \includegraphics[width=0.8\linewidth]{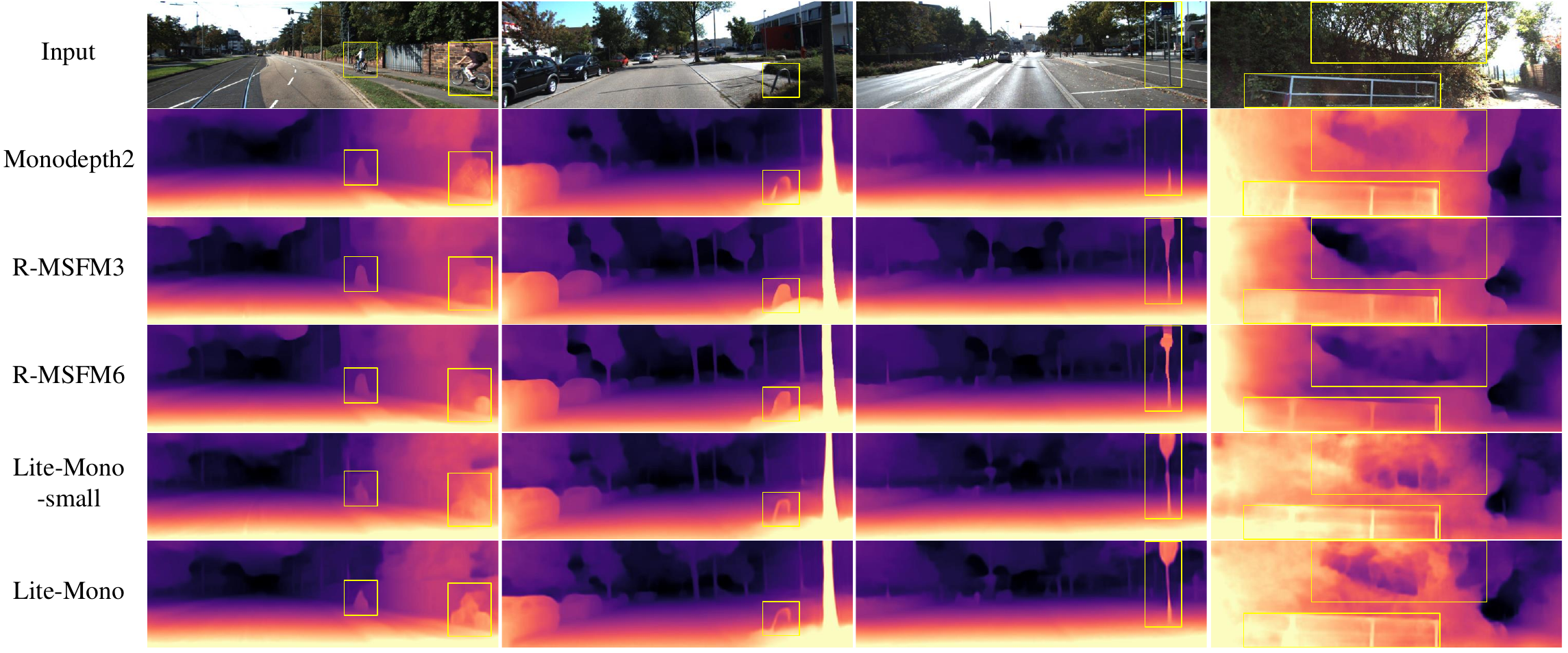}

   \caption{\textbf{Qualitative results on KITTI.} Here are some depth maps generated by Monodepth2~\cite{9}, R-MSFM3~\cite{10}, R-MSFM6~\cite{10}, Lite-Mono-small (ours), and Lite-Mono (ours), respectively. Monodepth2 and R-MSFM have limited receptive fields, so they yield some inaccurate depth predictions. Instead, our models can generate better results.}
   \label{fig:qualiti}
   \vspace{-2mm}
\end{figure*}
\subsection{Self-supervised learning}
\label{sec:ssl}
Different from the supervised training that utilizes ground truth of depth this work treats depth estimation as the task of image reconstruction. Similar to ~\cite{6} the learning objective is modeled to minimize an image reconstruction loss $\mathcal{L}_r$ between a target image $I_t$ and a synthesized target image $\hat{I_t}$, and an edge-aware smoothness loss $\mathcal{L}_{smooth}$ constrained on the predicted depth map $D_t$.

\textbf{Image reconstruction loss.}
The photometric reprojection loss is defined as:
\begin{equation}
 \mathcal{L}_p(\hat{I_t}, I_t) = \mathcal{L}_p(\mathcal{F}(I_s, P, D_t, K), I_t),
  \label{eq:lp}
\end{equation}
where $\hat{I_t}$ can be obtained by a function $\mathcal{F}$ in terms of the source image $I_s$, the estimated pose $P$, the predicted depth $D_t$, and the camera's intrinsics $K$. As introduced in~\cite{6} $\mathcal{L}_p$ is computed by a sum of the pixel-wise similarity $SSIM$ (Structural Similarity Index~\cite{22}) and the $L1$ loss between $\hat{I_t}$ and $I_t$:
\begin{equation}
 \mathcal{L}_p(\hat{I_t}, I_t) = \alpha \frac{1-SSIM(\hat{I_t}, I_t)}{2}+(1-\alpha)\lVert \hat{I_t}-I_t\rVert,
  \label{eq:prl}
\end{equation}
where $\alpha$ is set to 0.85 empirically~\cite{9}. In addition, to deal with out-of-view pixels and occluded objects in a source image the minimum photometric loss~\cite{9} is computed:
\begin{equation}
 \mathcal{L}_p(I_s, I_t) = \min\limits_{I_s\in [-1, 1]}\mathcal{L}_p(\hat{I_t}, I_t),
  \label{eq:minprl}
\end{equation}
where $I_s$ can be either the previous or the next frame with respect to the target image. Another binary mask~\cite{9} is used to remove moving pixels:
\begin{equation}
 \mu =\min\limits_{I_s\in [-1, 1]}\mathcal{L}_p(I_s, I_t) \textgreater \min\limits_{I_s\in [-1, 1]}\mathcal{L}_p(\hat{I_t}, I_t).
  \label{eq:miu}
\end{equation}
Therefore, the image reconstruction loss is defined as:
\begin{equation}
 \mathcal{L}_r(\hat{I_t}, I_t) =\mu \cdot\mathcal{L}_p(I_s, I_t),
  \label{eq:irl}
\end{equation}

\textbf{Edge-aware smoothness loss.} To smooth the generated inverse depth maps an edge-aware smoothness loss is calculated, followed by~\cite{9,14}:
\begin{equation}
 \mathcal{L}_{smooth} = \left|\partial_x\mathrm{d}_t^{\ast}\right|e^{-\left|\partial_xI_t\right|}+\left|\partial_x\mathrm{d}_t^{\ast}\right|e^{-\left|\partial_yI_t\right|},
  \label{eq:smooth}
\end{equation}
where $d_t^\ast = d_t/\hat{d_t}$ denotes the mean-normalized inverse depth. The total loss can be expressed as:
\begin{equation}
 \mathcal{L} = \frac{1}{3}\sum_{s\in \{1, \frac{1}{2}, \frac{1}{4}\}}(\mathcal{L}_r+\lambda \mathcal{L}_{smooth}),
  \label{eq:tl}
\end{equation}
where $s$ is the different scale output by the depth decoder. $\lambda$ is set to $1e^{-3}$ as in~\cite{9}.

\section{Experiments}
\label{sec:exp}
This section evaluates the proposed framework and demonstrates the superiority of Lite-Mono. 

\subsection{Datasets}
\textbf{KITTI.} The KITTI~\cite{29} dataset contains 61 stereo road scenes for research in autonomous driving and robotics, and it was collected by multiple sensors, including camera, 3D Lidar, GPU/IMU, etc. To train and evaluate the proposed method the Eigen split~\cite{30} is used, which has a total of 39,180 monocular triplets for training, 4,424 for evaluation, and 697 for testing. The self-supervised training is based on the known camera intrinsics $K$, as indicated in Eq.~\ref{eq:lp}. By averaging all the focal lengths of images across the KITTI dataset this paper uses the same intrinsics for all images during training~\cite{9}. In the evaluation the predicted depth is restricted in the range of $[0, 80]$m, as is common practice. 

\textbf{Make3D.} To evaluate the generalization ability of the proposed method it is further tested on the Make3D~\cite{31} dataset, which contains 134 test images of outdoor scenes. The model trained on the KITTI dataset is loaded and inferred directly on these test images.

\subsection{Implementation details}
\textbf{Hyperparameters.} The proposed method is implemented in PyTorch and trained on a single NVIDIA TITAN Xp with a batch size of 12. AdamW~\cite{24} is the optimizer, and the weight decay is set to $1e^{-2}$. Drop-path is used in the CDC and LGFI modules to mitigate overfitting. For models trained from scratch an initial learning rate of $5e^{-4}$ with a cosine learning rate schedule~\cite{23} is adopted, and the training epoch is set to 35. It is found that pre-training on ImageNet~\cite{36} makes the network converge fast, so the network is trained for 30 epochs when using pre-trained weights, and the initial learning rate is set to $1e^{-4}$. A monocular training for 35 epochs takes about 15 hours. 

\textbf{Data augmentation.} 
Data augmentation is adopted as a preprocessing step to improve the robustness of the training. To be specific, the following augmentations are performed with a $50\%$ chance: horizontal flips, brightness adjustment ($\pm 0.2$), saturation adjustment ($\pm 0.2$), contrast adjustment ($\pm 0.2$), and hue jitter ($\pm 0.1$). These adjustments are applied in a random order, and the same augmentation method is also used by~\cite{9,27, 10}. 

\textbf{Evaluation metrics} Accuracy is reported in terms of seven commonly used metrics proposed in~\cite{1}, which are Abs Rel, Sq Rel, RMSE, RMSE log, $\delta <1.25$, $\delta <1.25^2$, and $\delta <1.25^3$.

\subsection{KITTI results}
The proposed framework is compared with other representative methods with model sizes less than 35M, and the results are shown in Table~\ref{exp:full}. Lite-Mono beats all methods except MonoViT-tiny and is the smallest model (3.1M). Specifically, Lite-Mono greatly exceeds Monodepth2~\cite{9} with a ResNet18~\cite{12} backbone, but the model size is only about one-fifth of this model. It also outperforms the ResNet50 version of Monodepth2, which is the largest model (32.5M) in this table. Besides, Lite-Mono surpasses the recent well-designed small model R-MSFM~\cite{10}. Compared with the new MonoFormer~\cite{18} with a ResNet50 backbone the proposed Lite-Mono outperforms it in all metrics. Our other two smaller models also achieve satisfactory results, considering that they have fewer trainable parameters. In the last two rows of the table, the proposed Lite-Mono-8M also performs better than MonoViT-tiny, the smallest model of MonoViT~\cite{46}, with fewer parameters. Figure~\ref{fig:qualiti} shows that Lite-Mono achieves satisfactory results, even on challenging images where moving objects are close to the camera (column 1).

\subsection{Make3D results}
\begin{figure}[!thb]
  \centering
   \includegraphics[width=0.9\linewidth, height=3.5cm]{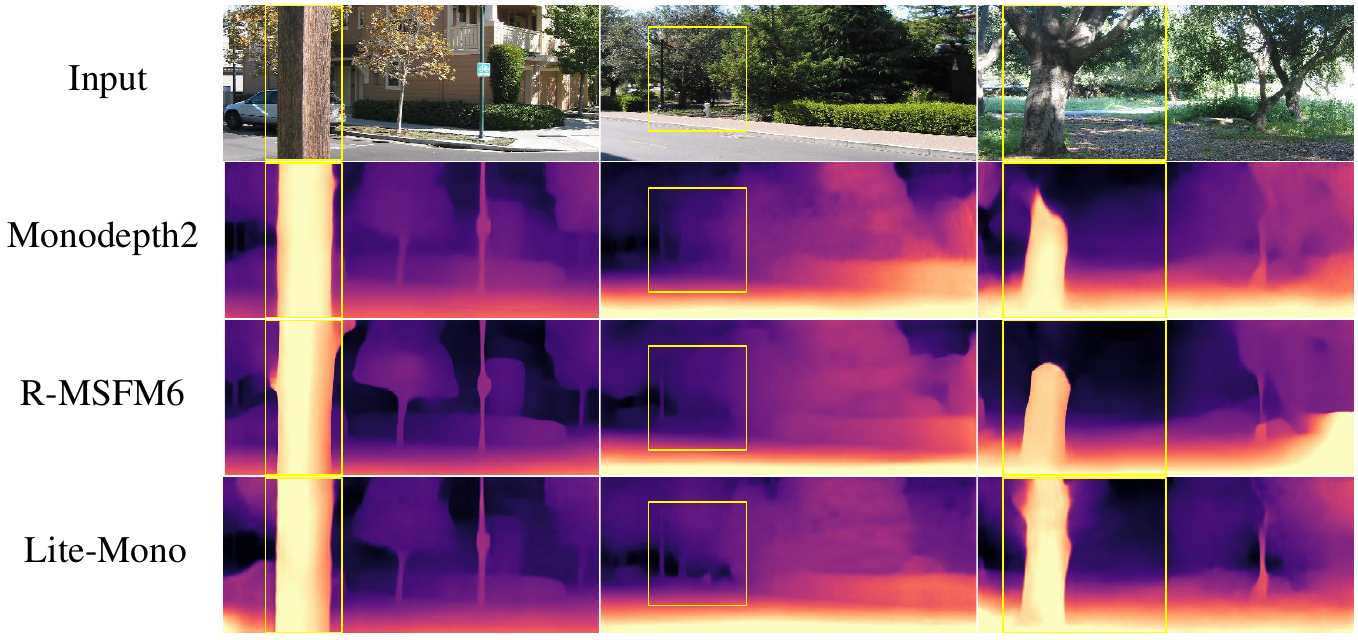}

   \caption{\textbf{Qualitative results on the Make3D dataset.} Lite-Mono is compared to Monodepth2~\cite{9} and R-MSFM~\cite{10}. Lite-Mono can perceive different sizes of objects.}
   \label{fig:make3d}
   \vspace{-3mm}
\end{figure}

The proposed method is evaluated on the Make3D dataset to show its generalization ability in different outdoor scenes. The model trained on KITTI is directly inferred without any fine-tuning. Table~\ref{exp:make3d} shows the comparison of Lite-Mono with the other three methods, and Lite-Mono performs the best. Figure~\ref{fig:make3d} shows some qualitative results. Owing to the proposed feature extraction modules Lite-Mono is able to model both local and global contexts, and perceives objects with different sizes.

\begin{table}[tbh]
\begin{center}
\scalebox{0.8}{
\begin{tabular}{c|cccc}
\hline
Method & Abs Rel & Sq Rel & RMSE & RMSE log \\
\hline
DDVO~\cite{25}&0.387&4.720&8.090&0.204\\
Monodepth2~\cite{9}&0.322&3.589&7.417&0.163\\
R-MSFM6~\cite{10}&0.334&3.285&7.212&0.169\\

Lite-Mono (Ours) &\textbf{0.305}  &   \textbf{3.060}  &   \textbf{6.981}  &   \textbf{0.158}\\
\hline
\end{tabular}}
\end{center}
\vspace{-4mm}
\caption{\textbf{Comparison of the proposed Lite-Mono to some other methods on the Make3D~\cite{31} dataset.} All models are trained on KITTI~\cite{29} with an image resolution of $640\times 192$.}
\label{exp:make3d}
\vspace{-4mm}
\end{table}

\subsection{Complexity and speed evaluation}
The proposed models' parameters, FLOPs (floating point of operations), and inference time are evaluated on an NVIDIA TITAN Xp and a Jetson Xavier and are compared with Monodepth2~\cite{9}, R-MSFM~\cite{10}, and MonoViT-tiny~\cite{46}. Table~\ref{exp:complexity} shows that the proposed design has a good balance between model size and speed. Notice that Lite-Mono-tiny outperforms Monodepth2 both in speed and accuracy (Table~\ref{exp:full}). Although R-MSFM~\cite{10} is a lightweight model it is slow. The latest MonoViT-tiny~\cite{46} runs the slowest due to its parallel blocks and multiple layers of self-attention. Our models also infer quickly on the Jetson Xavier,
which allows them to be used on edge devices.

\begin{table*}[!htb]
\begin{center}
\scalebox{0.7}{

\begin{tabular}{c|cc|cc|cc|cc}
\hline
&\multicolumn{2}{c|}{Encoder}&\multicolumn{2}{c|}{Decoder}&\multicolumn{2}{c|}{Full Model}&\multicolumn{2}{c}{Speed (ms)}\\
Method & Params. (M) & FLOPs (G) & Params. (M) & FLOPs (G) &Params. (M) & FLOPs (G) & Titan XP&Jetson Xavier\\
\hline
Monodepth2~\cite{9}&11.2&4.5&3.1&3.5&14.3&8.0&3.8&14.3\\

R-MSFM3~\cite{10}&0.7&2.4&2.8&14.1&3.5&16.5&7.8&22.3\\
R-MSFM6~\cite{10}&0.7&2.4&3.1&28.8&3.8&31.2&13.1&41.7\\
MonoViT-tiny~\cite{46}&5.6&7.8&4.7&15.9&10.3&23.7&13.5&47.4\\
\hline
Lite-Mono-tiny (Ours)&2.0&2.4&0.2&0.5&\textbf{2.2}&\textbf{2.9}&\textbf{3.3}&\textbf{12.7}\\
Lite-Mono-small (Outs)&2.3&4.1&0.2&0.7&2.5&4.8&4.3&19.2\\
Lite-Mono (Ours)&2.9&4.4&0.2&0.7&3.1&5.1&4.5&20.0\\
Lite-Mono-8M (Ours)&8.1&9.5&0.6&1.7&8.7&11.2&6.5&32.2\\

\hline

\end{tabular}}
\end{center}
\vspace{-4mm}
\caption{\textbf{Model complexity and speed evaluation.} We compare parameters, FLOPs (floating point of operations), and inference speed. The input size is $640\times 192$, and the batch size is 16.}
\label{exp:complexity}
\end{table*}

\begin{table*}[!htb]
\begin{center}
\scalebox{0.7}{

\begin{tabular}{c|c|c|cccc|ccc}
\hline
Architecture & Params. & Speed(ms) & Abs Rel & Sq Rel & RMSE & RMSE log &$\delta <1.25$ &$\delta <1.25^2$&$\delta <1.25^3$\\
\hline
Lite-Mono full model & 3.069M &4.5&\textbf{0.107}  &   \textbf{0.765}  &   \textbf{4.561}  &   \textbf{0.183}  &   \textbf{0.886}  &   \textbf{0.963}  &   \textbf{0.983}\\
\hline

w/o LGFI blocks&2.661M&3.2&   0.111  &   0.854  &   4.705  &   0.187  &   0.881  &   0.960  & 0.982\\

w/o dilated convolutions&3.069M&4.5&  0.112  &   0.836  &   4.685  &   0.187  &   0.880  &   0.960  &   0.982\\

w/o pooled concatenations&3.062M&4.4&0.109&0.842&4.700&0.186&0.883&0.960&0.982\\
w/o cross-stage connections&2.942M&4.4&0.108  &   0.834  &   4.683  &   0.185  &   0.884  &   0.962  &   0.982  \\
\hline

\end{tabular}}
\end{center}
\vspace{-4mm}
\caption{\textbf{Ablation study on model architectures.} All the models are trained and tested on KITTI with the input size $640\times 192$.}
\label{exp:ablation}
\vspace{-3mm}
\end{table*}

\subsection{Ablation study on model architectures}
To further demonstrate the effectiveness of the proposed model the ablation study is conducted to evaluate the importance of different designs in the architecture. We remove or adjust some modules in the network, and report their results on KITTI, as shown in Table~\ref{exp:ablation}. 

\textbf{The benefit of LGFI blocks.} When all the LGFI blocks in stage 2, 3, and 4 are removed, the model size decreases by 0.4M, but the accuracy also drops. The proposed LGFI is crucial to make Mono-Lite encode long-range global contexts, thus making up for the drawback that CNNs can only extract local features.

\textbf{The benefit of dilated convolutions.} If all the dilation rates of convolutions in CDC blocks are set to 1,~\textit{i.e.,} there are no dilated convolutions used in the network. It can be observed that although the model size remains the same the accuracy drops more than for not using LGFI blocks. The benefit of introducing the CDC module is to enhance the locality by gradually extracting multi-scale features, while not adding additional trainable parameters.

\textbf{The benefit of pooled concatenations.} Accuracy also decreases when three pooled concatenations are removed. This is because when using a down-sampling layer to reduce the size of feature maps, some spatial information is also lost. The advantage of using pooled concatenations is that the spatial information is kept, and this design only adds a small number of parameters (0.007M).

\textbf{The benefit of cross-stage connections.} When two cross-stage connections are removed the accuracy decreases slightly. The benefit of the proposed cross-stage connections in Mono-Lite is to promote feature propagation and cross-stage information fusion.

\subsection{Ablation study on dilation rates}
The influence of the dilation rate in the proposed CDC module on the accuracy is studied. Four different settings are used in this experiment.~\textbf{(1)} The default setting of our models is to group every three CDC blocks together, and set the dilation rate to 1, 2, and 3, respectively. For the last three blocks we set them to 2, 4, and 6. \textbf{(2)} Based on the default setting the dilation rates of the last three blocks are set to 1, 2, and 3, respectively. \textbf{(3)} Similar to the default setting we set 1, 2, and 5 in every three CDC blocks as a group. \textbf{(4)} Based on the default setting the dilation rates in the last two CDC blocks are set to 2, 4, 6 and 4, 8, 12, respectively. Table~\ref{exp:size} lists the accuracy under different dilation settings. Comparing~\textbf{(2)} with~\textbf{(3)} the accuracy benefits from larger dilation rates. However,~\textbf{(4)} using very large dilation rates in the late CDC blocks does not help. Simply pursuing larger dilation rates will result in the loss of local information, which is not good for the network to perceive small and medium-sized objects. Therefore, the proposed Lite-Mono adopts the setting~\textbf{(1)} to extract multi-scale local features,~\textit{i.e.,} smaller dilation rates are used in shallow layers, and they are doubled in the last three CDC modules.

\begin{table}[!htb]
\begin{center}
\scalebox{0.7}{

\begin{tabular}{c|cccc|ccc}
\hline
NO. & Abs Rel & Sq Rel & RMSE & RMSE log &$\delta_1$ &$\delta_2 $&$\delta_3$\\
\hline
1&\textbf{0.107}  &   \textbf{0.765}  &   \textbf{4.561}  &   \textbf{0.183}  &   \textbf{0.886}  &   \textbf{0.963}  &   \textbf{0.983}\\
2& 0.110  &   0.867  &   4.681  &   0.187  &   0.885  &   0.961  &   0.981\\
3& 0.108  &   0.835  &   4.652  &   0.186  &   0.885  &   0.962  &   0.982 \\
4& 0.110  &   0.855  &   4.642  &   0.187  &   0.885  &   0.961  &   0.981 \\

\hline

\end{tabular}}
\end{center}
\vspace{-5mm}
\caption{\textbf{Ablation study on dilation rates.}}
\label{exp:size}
\vspace{-4mm}
\end{table}

\section{Conclusions}
\label{sec:conclusions}
This paper presents a novel architecture Lite-Mono for lightweight self-supervised monocular depth estimation. A hybrid CNN and Transformer architecture is designed to model both multi-scale enhanced local features and long-range global contexts. The experimental results on the KITTI dataset demonstrate the superiority of our method. By setting optimized dilation rates in the proposed CDC blocks and inserting the LGFI modules to obtain the local-global feature correlations, Lite-Mono can perceive different scales of objects, even challenging moving objects closed to the camera. The generalization ability of the model is also validated on the Make3D dataset. Besides, Lite-Mono achieves a good trade-off between model complexity and inference speed. 

\section*{Acknowledgements}
This project has received funding from the European Union’s Horizon 2020 Research and Innovation Programme and the Korean Government under Grant Agreement No 833435. Content reflects only the authors’ view and the Research Executive Agency (REA) and the European Commission are not responsible for any use that may be made of the information it contains.

This work also uses the Geospatial Computing Platform of the Center of Expertise in Big Geodata Science (CRIB) (https://crib.utwente.nl). We thank Dr. Serkan Girgin for providing the computing infrastructure.



{\small
\bibliographystyle{ieee_fullname}
\bibliography{egbib}
}

\end{document}